\documentclass[sigconf]{acmart}

\AtBeginDocument{%
  \providecommand\BibTeX{{%
    \normalfont B\kern-0.5em{\scshape i\kern-0.25em b}\kern-0.8em\TeX}}}

\setcopyright{acmlicensed}
\copyrightyear{This is the preprint version of the paper accepted at the Fourth International Conference on AI ML Systems 2024. The final version will be published in the ACM proceedings.}
\acmYear{ }

\acmConference[AIMLSystems 2024]{The Fourth International Conference on AI ML Systems}{October 8-11, 2024}{Baton Rouge, USA}
\begin{document}

\title{Aggregated Knowledge Model: Enhancing Domain-Specific QA \\with Fine-Tuned and Retrieval-Augmented Generation Models}

\author{Fengchen Liu}
\email{fengchenliu@lbl.gov}
\orcid{0009-0006-4632-9604}
\affiliation{%
  \institution{Lawrence Berkeley National Laboratory}
  \streetaddress{1 Cyclotron Road}
  \city{Berkeley}
  \state{California}
  \country{USA}
  \postcode{94720}
}
\affiliation{%
  \institution{University of California, Berkeley}
  \streetaddress{University Avenue and, Oxford St}
  \city{Berkeley}
  \state{California}
  \country{USA}
  \postcode{94720}
}

\author{Jordan Jung}
\email{jordanjung@lbl.gov}
\affiliation{%
  \institution{Lawrence Berkeley National Laboratory}
  \streetaddress{1 Cyclotron Road}
  \city{Berkeley}
  \state{California}
  \country{USA}
  \postcode{94720}
}
\affiliation{%
  \institution{Convergent Computing/CCO}
  \streetaddress{1450 Maria Ln}
  \city{Walnut Creek}
  \state{California}
  \country{USA}
  \postcode{94596}
}

\author{Wei Feinstein}
\email{wfeinstein@lbl.gov}
\affiliation{%
  \institution{Lawrence Berkeley National Laboratory}
  \streetaddress{1 Cyclotron Road}
  \city{Berkeley}
  \state{California}
  \country{USA}
  \postcode{94720}
}

\author{Jeff D'Ambrogia}
\email{jeffd@lbl.gov}
\affiliation{%
  \institution{Lawrence Berkeley National Laboratory}
  \streetaddress{1 Cyclotron Road}
  \city{Berkeley}
  \state{California}
  \country{USA}
  \postcode{94720}
}

\author{Gary Jung}
\email{gmjung@lbl.gov}
\affiliation{%
  \institution{Lawrence Berkeley National Laboratory}
  \streetaddress{1 Cyclotron Road}
  \city{Berkeley}
  \state{California}
  \country{USA}
  \postcode{94720}
}

\renewcommand{\shortauthors}{Liu and Jung, et al.}


\begin{abstract}
This paper introduces a novel approach to enhancing closed-domain Question Answering (QA) systems, focusing on the specific needs of the Lawrence Berkeley National Laboratory (LBL) Science Information Technology (ScienceIT) domain. Utilizing a rich dataset derived from the ScienceIT documentation, our study embarks on a detailed comparison of two fine-tuned large language models and five retrieval-augmented generation (RAG) models. Through data processing techniques, we transform the documentation into structured context-question-answer triples, leveraging the latest Large Language Models (AWS Bedrock, GCP PaLM2, Meta LLaMA2, OpenAI GPT-4, Google Gemini-Pro) for data-driven insights. Additionally, we introduce the Aggregated Knowledge Model (AKM), which synthesizes responses from the seven models mentioned above using K-means clustering to select the most representative answers. The evaluation of these models across multiple metrics offers a comprehensive look into their effectiveness and suitability for the LBL ScienceIT environment. The results demonstrate the potential benefits of integrating fine-tuning and retrieval-augmented strategies, highlighting significant performance improvements achieved with the AKM. The insights gained from this study can be applied to develop specialized QA systems tailored to specific domains.
\end{abstract}



\maketitle

\section{Introduction}

Rapid advancements in Natural Language Processing (NLP) and deep learning have enhanced the capability of Question Answering (QA) systems to utilize both structured and unstructured data effectively. These systems are increasingly critical in diverse domains, from customer support to academic research, particularly in closed-domain applications known for their precision and domain-specific focus.

At the Lawrence Berkeley National Laboratory (LBL), the ScienceIT department addresses the IT needs of the scientific community, utilizing extensive documentation and resources. This work leverages data from the LBL ScienceIT documentation website, processed to support our QA system. By integrating technologies like Google Cloud VertexAI, AWS Bedrock, OpenAI, and Meta LLaMA with Retrieval Augmented Generation (RAG) \cite{lewis2020retrieval, guu2020retrieval} through the LangChain framework, we fine-tune multiple models for enhanced QA performance. In addition to these models, we propose the Aggregated Knowledge Model (AKM), which synthesizes responses from the various fine-tuned and RAG models using K-means clustering \cite{hartigan1979algorithm, hamerly2003learning} to select the most representative answers. This novel approach aims to improve the overall performance and reliability of QA systems in the LBL ScienceIT domain.

This paper presents a comparative study of seven models, analyzing their performance and potential in the LBL ScienceIT domain. Additionally, the Aggregated Knowledge Model (AKM) is introduced and evaluated. The findings from this study offer insights into the development of closed-domain QA systems and suggest potential directions for future domain-specific QA enhancements.

\subsection{Motivation and Benefits}
\begin{itemize}
\item Streamlining Information Retrieval for Researchers: 
The LBL researchers require efficient access to ScienceIT documentation to support their work, necessitating systems that provide quick, precise information retrieval.

\item  Elevating User Experience through Accessible Expertise: 
The QA system enhances user experience by providing rapid and accurate responses, improving efficiency and focus on primary research tasks.

\item  Sustainability in Knowledge Dissemination: 
Our QA system offers a sustainable solution for information dissemination, evolving with the expanding needs of the ScienceIT ecosystem.

\item  Bridging Informational Gaps: 
The initiative promotes accessible, accurate information for all researchers, ensuring inclusivity and comprehensive knowledge access within the Berkeley Lab.
\end{itemize}

\section{Related Work}
The evolution of QA systems began with Stanford's BASEBALL in 1961, a rule-based linguistic model interfaced with a database \cite{green1961baseball}. Advancements in deep learning have significantly impacted QA system development across various domains, notably through the use of neural networks like CNNs \cite{lei2018novel}, LSTMs \cite{xia2018novel}, and the transformational architecture of transformers such as BERT and GPT-2 \cite{vaswani2017attention, devlin2018bert, ethayarajh2019contextual}. Modern QA approaches include:
\begin{itemize}
\item  Information Retrieval QA, utilizing search engines for answer retrieval \cite{azad2019query}.
\item  NLP QA, employing NLP and machine learning for response deduction \cite{garg2019automating}.
\item  Knowledge Base QA, using structured datasets for precise answers \cite{zhou2021dfm}.
\item  Hybrid QA, combining multiple methods for enhanced results \cite{loginova2021towards}.
\item  Retrieval-Augmented Generation QA, integrating retrieval with generative methods for contextually accurate responses \cite{lewis2020retrieval, guu2020retrieval}.
\item  Open Domain and Closed Domain QA, catering to general or specific informational needs \cite{kia2022adaptable,alzubi2021cobert}.
\end{itemize}
This diversity highlights the evolution from singular, rule-based models to complex, multifaceted systems that blend retrieval, NLP, and domain-specific methodologies.

\section{Data}
\label{sec:data}
Generating meaningful question-answer pairs for a closed-domain QA system presents a significant challenge, particularly in ensuring the relevance and accuracy of the pairs to the specific domain. To address this challenge, we turned to the capabilities of the Language Model (LLM). Specifically, we employed GCP VertexAI's PaLM2(chat-bison@001)\cite{Google_2023a}. Given a context paragraph, this LLM was tasked with producing concise and pertinent question-answer pairs. The model's ability to understand the essence of a paragraph and distill it into questions and answers was crucial in meeting this challenge. The depth and breadth of the context paragraph determined the number of question-answer pairs generated, with an average context yielding about 5 pairs. In total, our approach resulted in approximately 2800 tuples of context, question, and answer for our data-driven experiments.

\section{Models}
In the domain of Question Answering (QA), the choice of the underlying model plays a pivotal role in determining the system's efficiency and accuracy. Given the specific requirements of the LBL ScienceIT domain, we ventured into a comprehensive exploration of both conventional and contemporary models. Our endeavor was twofold: fine-tuning Large Language Models (LLMs) and harnessing the power of Retrieval Augmented Generation (RAG) models. This section delineates the specifics of each model we employed.

\begin{figure*}[h]
    \centering
    \includegraphics[width=1\textwidth]{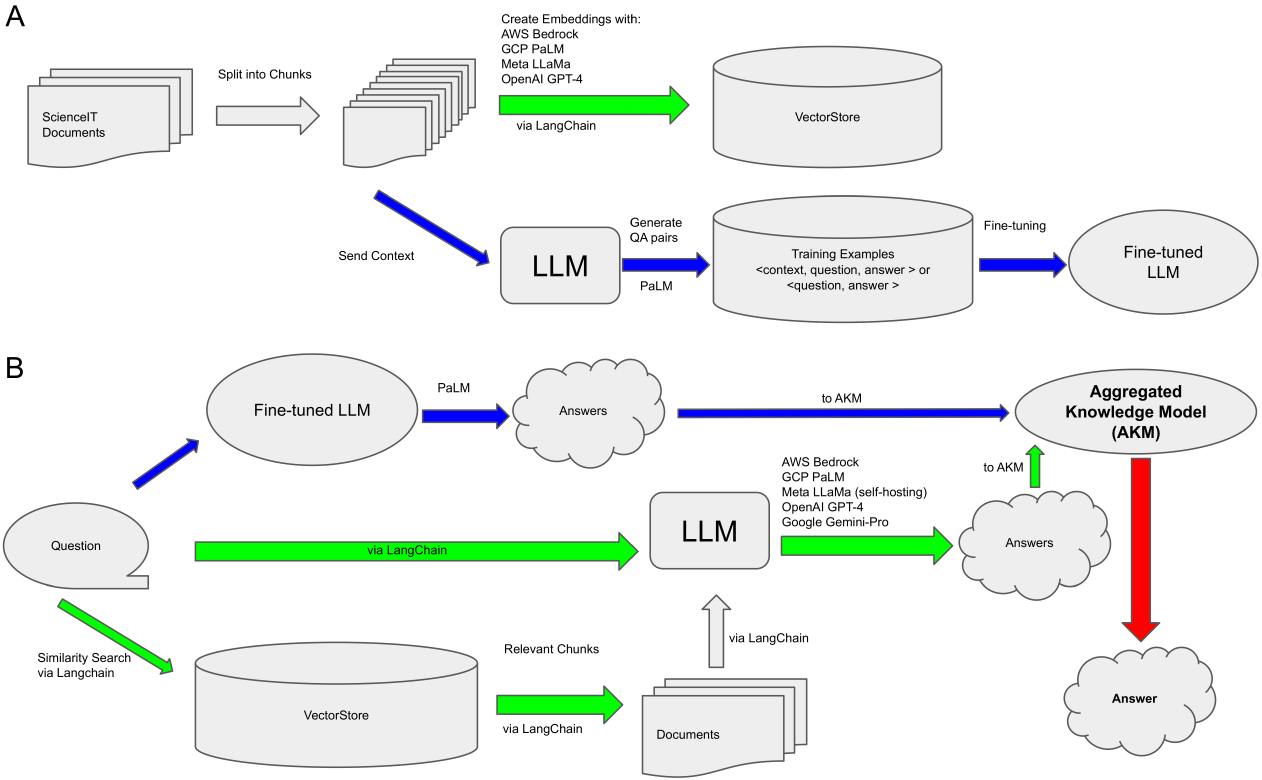} 
    \caption{Workflow of the Enhanced ScienceIT QA System. This diagram illustrates the two-fold process in the QA system's development. Panel A shows the data preparation phase, where ScienceIT documents are split into chunks and processed to generate embeddings with AWS Bedrock, GCP PaLM, Meta LLaMA, and OpenAI GPT-4, facilitated by LangChain, creating a VectorStore for RAG (green) and training examples for model fine-tuning (blue). Panel B depicts the query processing flow where a fine-tuned LLM answers a user’s question, and the RAG retrieves relevant document chunks from the VectorStore using similarity search. The question, along with retrieved chunks, is passed through various LLMs, including AWS Bedrock, GCP PaLM, Meta LLaMA (self-hosted), OpenAI GPT-4, and Google Gemini-Pro, to generate precise answers. The Aggregated Knowledge Model (AKM) further enhances the system by synthesizing answers from the fine-tuned and RAG models using K-means clustering to select the most representative answer (red), improving overall accuracy and reliability.}
    \label{fig:workflow}
\end{figure*}

Figure[\ref{fig:workflow}] shows a systematic workflow to tackle QA challenges. The process begins with document preparation which involves loading the data, fragmenting it into digestible segments (discussed in Section \ref{sec:data}), and then encoding these segments into numerical vectors using the AWS Bedrock Titan \cite{AWS-Bedrock_2023}, GCP PaLM2 \cite{Google_2023a}, Meta LLaMA \cite{touvron2307llama} and OpenAI GPT-4 \cite{achiam2023gpt} embeddings. Once transformed, as shown in Figure[\ref{fig:workflow}]A, these vectors are cataloged in an index to streamline future retrievals. When a query is presented, as shown in Figure[\ref{fig:workflow}]B, its embedding is generated and juxtaposed with the indexed embeddings to pinpoint the most pertinent document segments. These identified segments are then incorporated into the context of the prompt dispatched to the AWS Bedrock, GCP PaLM2 (text-bison-001), the self-hosting Meta LLaMA models, OpenAI GPT-4 and Google Gemini-Pro\cite{team2023gemini}. This ensures that the response is not only relevant but also deeply contextual, drawing directly from the information within the retrieved document segments.

\subsection{The Enhanced ScienceIT QA System}

\subsubsection{Two fine-tuned models:}
\begin{itemize}
\item Model 1: Fine-tuned on \( \langle \text{question}, \text{answer} \rangle \) pairs.
   \[
   \text{Model}_1: \{(q_i, a_i)\} \rightarrow \hat{a}_i
   \]
   Where \( q_i \) is the question, \( a_i \) is the ground truth answer, and \( \hat{a}_i \) is the predicted answer.
\item Model 2: Fine-tuned on \( \langle \text{question}, \text{context}, \text{answer} \rangle \) triples.
   \[
   \text{Model}_2: \{(q_i, c_i, a_i)\} \rightarrow \hat{a}_i
   \]
   Where \( c_i \) is the context.
\end{itemize}
Both models were fine-tuned using 4x NVIDIA A100 80GB GPUs.

\subsubsection{Five RAG models:}
\begin{itemize}
\item Embedding Generation: For each document segment \( d_j \), generate an embedding \( \mathbf{e}_j \) using the respective LLM’s embedding model.
   \[
   \mathbf{e}_j = \text{Embed}(d_j)
   \]
\item Vector Store Creation: 
    Store all embeddings \( \mathbf{E} = \{\mathbf{e}_1, \mathbf{e}_2, \ldots, \mathbf{e}_n\} \) in a vector database.
\item Query Processing:
    Generate an embedding for the query \( q_i \):
     \[
     \mathbf{e}_{q_i} = \text{Embed}(q_i)
     \]
    Retrieve the top-k relevant document segments based on similarity (e.g., semantic similarity) to the query embedding:
     \[
     \{d_{i1}, d_{i2}, \ldots, d_{ik}\} = \text{Retrieve}(\mathbf{e}_{q_i}, \mathbf{E})
     \]
    Concatenate these retrieved segments to form the augmented context \( c'_i \).
\item Answer Generation:
   Generate the answer \( \hat{a}_i \) using the LLM with the augmented context:
     \[
     \hat{a}_i = \text{LLM}(q_i, c'_i)
     \]
\end{itemize}

The fifth model, which is a RAG model using Meta LLaMA2, was hosted using 4x NVIDIA A40 40GB GPUs.

\subsection{Aggregated Knowledge Model (AKM)}
To enhance our Domain Knowledge Question Answering system, we introduced the Aggregated Knowledge Model (AKM) as the 8th model. This model leverages the strengths of seven individual LLM models, including fine-tuned versions of Google PaLM2 and various Retrieval Augmented Generation (RAG) models from AWS Bedrock, GCP PaLM2, Meta LLaMA2, OpenAI GPT-4, and Google Gemini-Pro.

The AKM evaluates responses from these models by utilizing K-means clustering \cite{hartigan1979algorithm, hamerly2003learning} with \( k=1 \) to identify the most representative answer for each question. The specific steps involved are as follows:

\subsubsection{Vectorization}
Each predicted answer \( A_i \) from the models is converted into a TF-IDF \cite{ramos2003using} vector \( \mathbf{v}_i \):

\[
\mathbf{V} = [\mathbf{v}_1, \mathbf{v}_2, \ldots, \mathbf{v}_7]
\]

\subsubsection{Clustering}
We apply K-means clustering with \( k=1 \) to these vectors to find the centroid \( \mathbf{c} \) of the cluster:

\[
\mathbf{c} = \frac{1}{7} \sum_{i=1}^{7} \mathbf{v}_i
\]

\subsubsection{Distance Calculation}
For each vector \( \mathbf{v}_i \), we calculate its Euclidean distance to the centroid \( \mathbf{c} \):

\[
d_i = \|\mathbf{v}_i - \mathbf{c}\|
\]

\subsubsection{Selection}
The answer corresponding to the vector with the smallest distance to the centroid is selected as the final response:

\[
A_{\text{final}} = A_{\arg\min_{i} d_i}
\]

This approach ensures that the chosen answer is the most central and representative of the combined knowledge from all models. By synthesizing diverse perspectives and reducing the impact of outlier responses, the AKM provides more accurate and reliable answers.

\section{Experiments}

The primary goal of our experiments was to assess the efficacy of various model architectures for our Closed Domain QA System. Approximately 2800 (context, question, answer) tuples were used, with 80\% allocated for model fine-tuning and 20\% for validation to ensure thorough learning and unbiased assessment.

Evaluation was conducted using a testing dataset comprising approximately 560 ScienceIT domain knowledge questions. Each question was processed by the first seven models (two fine-tuned models and five RAG models) with the LLM temperature set to 0.5. The eighth model, AKM, aggregated the responses from these seven models to generate its answer. This entire process was repeated 100 times, resulting in a total of 56000 samples for evaluation. The performance metrics for each model were computed as the mean and standard deviation (in Table[\ref{tab:results}], and Appendix \ref{sec:appendix}) from these samples, ensuring robust and reliable results.

\subsection{Evaluation Metrics}
We utilized BLEU Scores to evaluate n-gram accuracy, ROUGE Scores to assess recall, precision, and F1 metrics, and STS (Semantic Textual Similarity) to examine the semantic similarity between model-generated and reference answers, emphasizing STS's relevance due to language variability \cite{post2018call}.

\subsection{Experimental Comparison}
In addition to comparing the performance of all eight models, we also experimented with several methods for the eighth model (AKM) to select the most representative answer:

\subsubsection{TF-IDF and Cosine Similarity}
We first vectorized the answers using TF-IDF \cite{ramos2003using} and computed the cosine similarity between each pair of answers. However, this method did not yield the best performance, as the cosine similarity metric alone was insufficient to capture the nuanced differences between the answers.

\subsubsection{Embeddings and Mean Embedding (using BERT)}
In another approach, we utilized pre-trained BERT embeddings \cite{devlin2018bert} to represent each answer. We then computed the mean embedding of all answers and selected the answer whose embedding was closest to this mean. Although this method showed improvement over cosine similarity, it still fell short in accurately identifying the most representative answers due to the high dimensionality and complexity of the embeddings.

\subsubsection{Clustering (using K-means)}
The most effective method was K-means clustering \cite{hartigan1979algorithm, hamerly2003learning}. By clustering the TF-IDF \cite{ramos2003using} vectors of the predicted answers and selecting the answer closest to the centroid, we were able to consistently identify the most representative answer. This method outperformed the previous approaches by better capturing the central tendency of the answer distributions.

\section{Analysis}
Our diverse selection of models yielded varied results, with significant insights into their performance across different metrics, as showcased in Figure[\ref{fig:results}] and Table[\ref{tab:results}] (for each model's metrics distribution, please see Appendix \ref{sec:appendix} Figure \ref{fig:distribution}). BLEU and ROUGE scores indicated specific strengths in text alignment and recall capabilities, while STS scores highlighted semantic similarities effectively handled by our models. Notably, Models with Retrieval-Augmented Generation (RAG) features, such as Model 6 (OpenAI GPT-4) and Model 7 (Google Gemini-Pro), showed strong performances in semantic contexts, aligning closely with reference answers, showcasing the impact of RAG on model performance.

The introduction of the Aggregated Knowledge Model (AKM) further improved performance by effectively synthesizing the strengths of the individual models. From the evaluation metrics in Table[\ref{tab:results}], it is evident that the AKM model consistently achieved higher scores across various metrics, indicating its robustness and effectiveness in providing accurate and contextually relevant answers. This improvement, with an overall performance increase of over 8\%, can be attributed to the AKM's ability to aggregate and balance the diverse responses from multiple models, capturing the central tendency and reducing the impact of outlier predictions.

The use of K-means clustering in the AKM proved to be particularly effective. By vectorizing the predicted answers and identifying the centroid of the cluster, the AKM could select the most representative answer for each question. This method outperformed others, such as TF-IDF with cosine similarity and BERT embeddings with mean embedding, by better capturing the overall distribution of the answers and selecting the one closest to the centroid.

\begin{figure*}[h]
    \centering
    \includegraphics[width=1\textwidth]{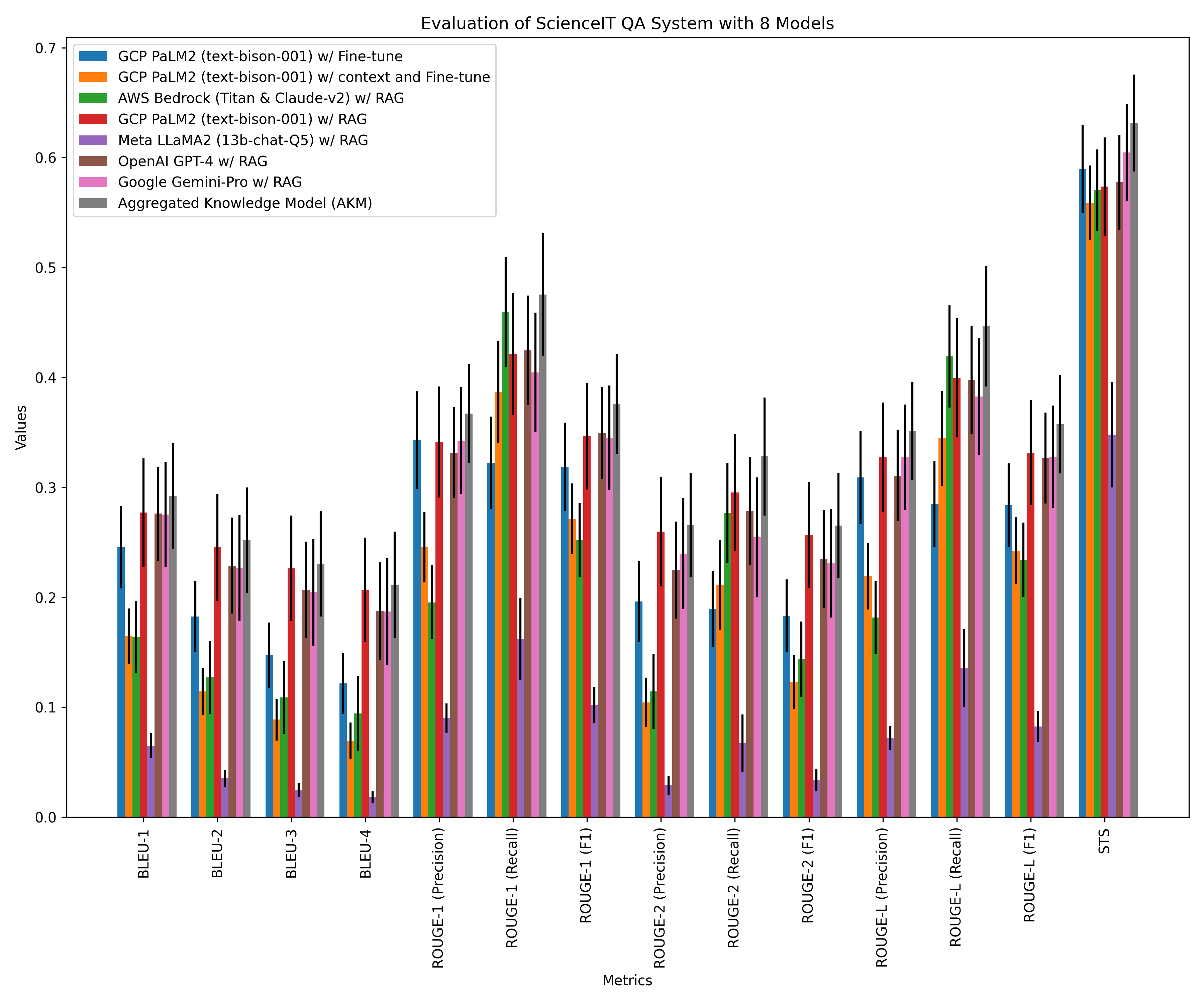} 
    \caption{Performance metrics of different models across various evaluation metrics. The figure compares eight models: two fine-tuned, five retrieval-augmented generation (RAG), and the Aggregated Knowledge Model (AKM). The models were evaluated using approximately 560 ScienceIT domain knowledge questions, with the process repeated 100 times. Metrics include BLEU-1, BLEU-2, BLEU-3, BLEU-4, ROUGE-1 (Precision), ROUGE-1 (Recall), ROUGE-1 (F1), ROUGE-2 (Precision), ROUGE-2 (Recall), ROUGE-2 (F1), ROUGE-L (Precision), ROUGE-L (Recall), ROUGE-L (F1), and Semantic Textual Similarity (STS). Each bar represents the mean value of the metric for a model, with error bars indicating the standard deviation. The AKM model aggregates responses from the seven models using K-means clustering, showing improved performance across the metrics.}

    \label{fig:results}
\end{figure*}

\begin{table*}
\centering
\small 
\begin{tabular}{l|cccccccc}
\hline
Metrics & Model 1 & Model 2 & Model 3 & Model 4 & Model 5 & Model 6 & Model 7 & Model 8 \\
\hline
BLEU-1 & 0.246 ± 0.038 & 0.165 ± 0.025 & 0.164 ± 0.033 & 0.277 ± 0.049 & 0.065 ± 0.011 & 0.276 ± 0.043 & 0.275 ± 0.048 & 0.292 ± 0.048 \\
BLEU-2 & 0.182 ± 0.032 & 0.115 ± 0.021 & 0.127 ± 0.033 & 0.246 ± 0.049 & 0.035 ± 0.008 & 0.229 ± 0.044 & 0.227 ± 0.048 & 0.252 ± 0.048 \\
BLEU-3 & 0.147 ± 0.030 & 0.089 ± 0.019 & 0.109 ± 0.034 & 0.226 ± 0.048 & 0.025 ± 0.006 & 0.207 ± 0.044 & 0.205 ± 0.048 & 0.231 ± 0.048 \\
BLEU-4 & 0.122 ± 0.028 & 0.070 ± 0.017 & 0.094 ± 0.034 & 0.207 ± 0.048 & 0.018 ± 0.005 & 0.188 ± 0.044 & 0.187 ± 0.049 & 0.211 ± 0.048 \\
ROUGE-1 (Precision) & 0.343 ± 0.045 & 0.245 ± 0.032 & 0.195 ± 0.034 & 0.341 ± 0.051 & 0.090 ± 0.013 & 0.332 ± 0.041 & 0.342 ± 0.049 & 0.367 ± 0.045 \\
ROUGE-1 (Recall) & 0.323 ± 0.042 & 0.387 ± 0.047 & 0.460 ± 0.050 & 0.422 ± 0.056 & 0.162 ± 0.038 & 0.425 ± 0.050 & 0.404 ± 0.054 & 0.476 ± 0.056 \\
ROUGE-1 (F1) & 0.319 ± 0.040 & 0.271 ± 0.032 & 0.252 ± 0.034 & 0.346 ± 0.048 & 0.102 ± 0.016 & 0.350 ± 0.042 & 0.345 ± 0.048 & 0.376 ± 0.045 \\
ROUGE-2 (Precision) & 0.196 ± 0.037 & 0.104 ± 0.023 & 0.114 ± 0.034 & 0.260 ± 0.050 & 0.029 ± 0.009 & 0.225 ± 0.044 & 0.240 ± 0.051 & 0.266 ± 0.048 \\
ROUGE-2 (Recall) & 0.189 ± 0.034 & 0.211 ± 0.041 & 0.277 ± 0.046 & 0.295 ± 0.053 & 0.067 ± 0.026 & 0.278 ± 0.049 & 0.255 ± 0.054 & 0.328 ± 0.054 \\
ROUGE-2 (F1) & 0.183 ± 0.033 & 0.123 ± 0.025 & 0.144 ± 0.034 & 0.257 ± 0.048 & 0.034 ± 0.010 & 0.235 ± 0.045 & 0.231 ± 0.049 & 0.265 ± 0.048 \\
ROUGE-L (Precision) & 0.309 ± 0.042 & 0.219 ± 0.030 & 0.181 ± 0.033 & 0.327 ± 0.050 & 0.072 ± 0.011 & 0.311 ± 0.041 & 0.327 ± 0.048 & 0.352 ± 0.044 \\
ROUGE-L (Recall) & 0.285 ± 0.039 & 0.345 ± 0.043 & 0.419 ± 0.047 & 0.400 ± 0.054 & 0.136 ± 0.035 & 0.398 ± 0.049 & 0.383 ± 0.053 & 0.447 ± 0.055 \\
ROUGE-L (F1) & 0.284 ± 0.038 & 0.243 ± 0.030 & 0.234 ± 0.034 & 0.332 ± 0.048 & 0.082 ± 0.014 & 0.327 ± 0.041 & 0.328 ± 0.047 & 0.357 ± 0.045 \\
STS & 0.590 ± 0.040 & 0.559 ± 0.034 & 0.570 ± 0.037 & 0.574 ± 0.045 & 0.348 ± 0.048 & 0.578 ± 0.043 & 0.605 ± 0.044 & 0.631 ± 0.044 \\
Average & 0.266 ± 0.037 & 0.225 ± 0.030 & 0.239 ± 0.037 & 0.322 ± 0.050 & 0.090 ± 0.018 & 0.311 ± 0.044 & 0.311 ± 0.049 & 0.347 ± 0.048 \\
\hline
\end{tabular}
\caption{Performance metrics (the mean and standard deviation (±)) across different models. Model 1: GCP PaLM2 (text-bison-001) w/ Fine-tune. Model 2: GCP PaLM2 (text-bison-001) w/ context and Fine-tune. Model 3: AWS Bedrock (Titan \& Claude-v2) w/ RAG. Model 4: GCP PaLM2 (text-bison-001) w/ RAG. Model 5: Meta LLaMA2 (13b-chat-Q5) w/ RAG. Model 6: OpenAI GPT-4 w/ RAG. Model 7: Google Gemini-Pro w/ RAG. Model 8: Aggregated Knowledge Model (AKM). STS: Semantic Textual Similarity.}
\label{tab:results}
\end{table*}

\section{Conclusion}

This study explored various models in Closed Domain Question Answering (QA) Systems, demonstrating the distinct advantages of Retrieval Augmented Generation models. The introduction of the Aggregated Knowledge Model (AKM) marked a significant improvement, with an overall performance increase of over 8\%, surpassing the previously best-performing OpenAI GPT-4 with RAG. The AKM model leveraged the strengths of multiple models to provide more accurate and reliable answers, resulting in a higher average evaluation score across multiple metrics.

The results emphasize the need for tailoring model architecture and fine-tuning strategies to specific domains. The AKM model, by aggregating responses from various LLMs and RAG models, highlights the importance of integrating diverse model capabilities and effective retrieval mechanisms to enhance performance.

Challenges persist in creating question-answer pairs for extensive document sets, necessitating innovative approaches and broader experiments to refine our models and outcomes. Future research will focus on enhancing data quality through advanced augmentation techniques, developing hybrid models that integrate various architectural strengths, and employing advanced metrics like the ReCOGS Semantic Exact Match \cite{wu2023recogs} for deeper semantic analysis. Additionally, addressing ethical implications, particularly around data privacy, user consent, and bias mitigation, is crucial as we advance these technologies.

\begin{acks}
This research used the Lawrencium computational cluster provided by the IT Division at Lawrence Berkeley National Laboratory, supported by the U.S. Department of Energy under Contract No. DE-AC02-05CH11231. We also acknowledge the support from the Tuition Assistance Program (TAP) at Lawrence Berkeley National Laboratory. This research was further supported by the Master of Information and Data Science program at the UC Berkeley School of Information, with guidance from its faculty and administration. Additionally, we acknowledge the computational resources from Amazon Web Services (AWS) and Google Cloud Platform (GCP) that supported our experiments.
\end{acks}

\bibliographystyle{Reference-Format}
\bibliography{lit-review}

\appendix
\section{Distribution of metrics}
\label{sec:appendix}

\begin{figure*}[h]
    \centering
    \includegraphics[width=1\textwidth]{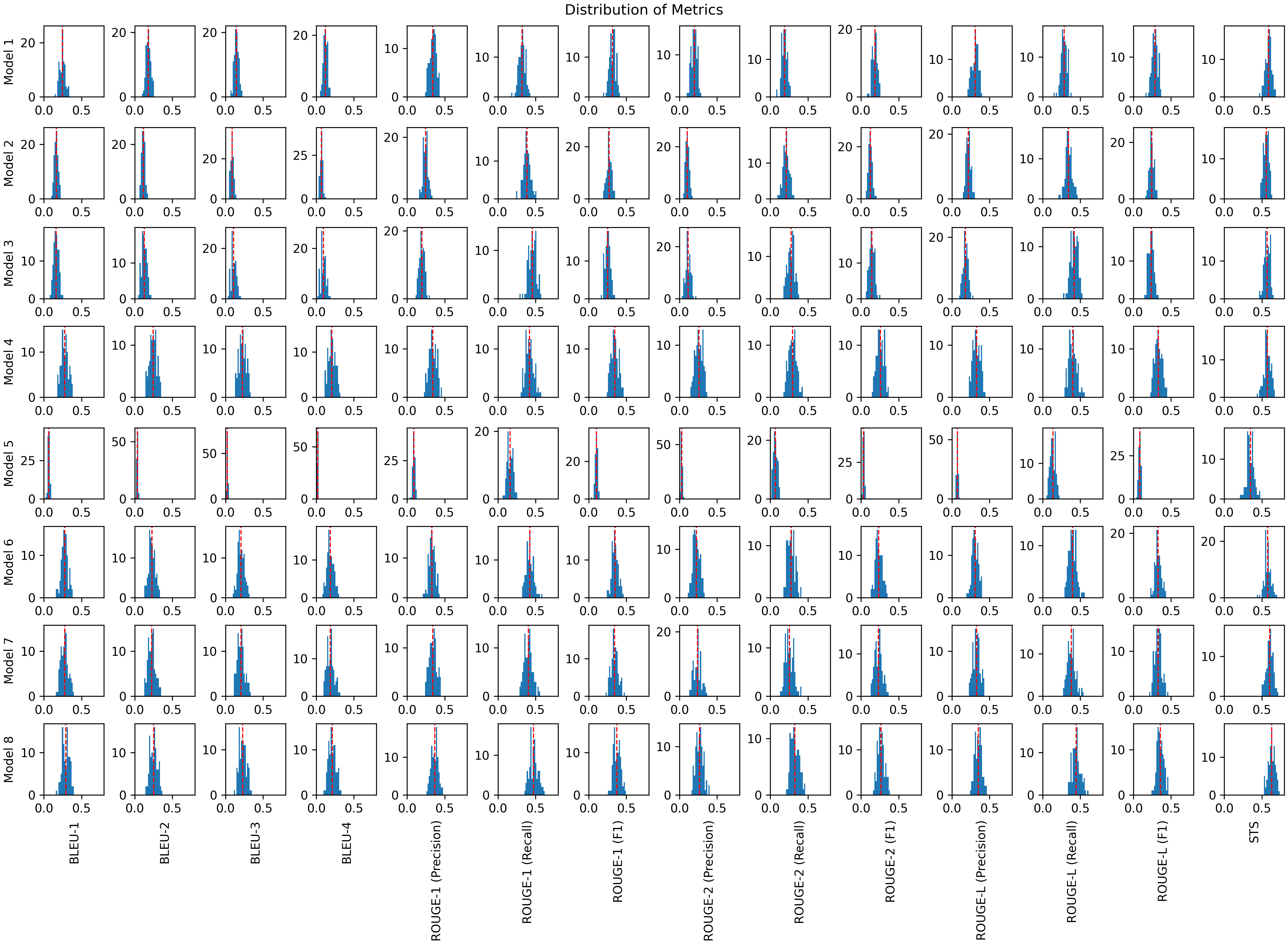} 
    \caption{Distribution of metrics for each model. The plot shows the performance metrics (BLEU, ROUGE, and STS) distributions for eight models evaluated on approximately 560 ScienceIT domain knowledge questions. Each model's performance was assessed 100 times, resulting in distributions. The red lines indicate the mean values. Model 1: GCP PaLM2 (text-bison-001) w/ Fine-tune. Model 2: GCP PaLM2 (text-bison-001) w/ context and Fine-tune. Model 3: AWS Bedrock (Titan \& Claude-v2) w/ RAG. Model 4: GCP PaLM2 (text-bison-001) w/ RAG. Model 5: Meta LLaMA2 (13b-chat-Q5) w/ RAG. Model 6: OpenAI GPT-4 w/ RAG. Model 7: Google Gemini-Pro w/ RAG. Model 8: Aggregated Knowledge Model (AKM). STS: Semantic Textual Similarity.}
    \label{fig:distribution}
\end{figure*}

\end{document}